\documentclass[10pt,a4paper,twocolumn]{article}

\usepackage{framed,multirow}
\usepackage{graphicx}

\usepackage{amssymb}
\usepackage{latexsym}
\usepackage{amsmath,amssymb,amsfonts}
\usepackage{url}
\usepackage{xcolor}

\usepackage{xurl}
\usepackage{hyperref}
\usepackage{threeparttable}
\usepackage{booktabs}
\usepackage{tabularx} % Used for adaptive table width
\usepackage[authoryear]{natbib} % Support citep

\usepackage{lmodern} % Font rendering 
\usepackage[T1]{fontenc}

\usepackage{authblk}
\usepackage{makecell}
\usepackage{adjustbox}

\definecolor{newcolor}{rgb}{.8,.349,.1}

\title{\textbf{Cross-modal ultra-scale learning with tri-modalities of renal biopsy images for glomerular multi-disease auxiliary diagnosis}}

\author[1]{\textbf{Kaixing Long}}
\author[1,2,3]{\textbf{Danyi Weng}}
\author[1,2,3]{\textbf{Yun Mi}}
\author[1,2,3]{\textbf{Zhentai Zhang}}
\author[4]{\textbf{Yanmeng Lu}}
\author[4]{\textbf{Zhitao Zhou}}
\author[5,6]{\textbf{Jian Geng}}
\author[1,2,3]{\textbf{Liming Zhong}}
\author[1,2,3]{\textbf{Qianjin Feng}}
\author[1,2,3]{\textbf{Wei Yang}}
\author[1,2,3]{\textbf{Lei Cao}\thanks{Corresponding author: \texttt{caolei@smu.edu.cn}}}

\affil[1]{School of Biomedical Engineering, Southern Medical University, Guangzhou, 510515, China}
\affil[2]{Guangdong Provincial Key Laboratory of Medical Image Processing, Guangzhou, 510515, China}
\affil[3]{Guangdong Province Engineering Laboratory for Medical Imaging and Diagnostic Technology, Guangzhou, 510515, China}
\affil[4]{Central Laboratory, Southern Medical University, Guangzhou 510515, China}
\affil[5]{Department of Pathology, School of Basic Medical Sciences, Southern Medical University, Guangzhou, 510515, China}
\affil[6]{Guangzhou Huayin Medical Laboratory Center, Guangzhou, 510515, China}

\date{}

\begin{document}
\maketitle

\begin{abstract}
Constructing a multi-modal automatic classification model based on three types of renal biopsy images can assist pathologists in glomerular multi-disease identification. However, the substantial scale difference between transmission electron microscopy (TEM) image features at the nanoscale and optical microscopy (OM) or immunofluorescence microscopy (IM) images at the microscale poses a challenge for existing multi-modal and multi-scale models in achieving effective feature fusion and improving classification accuracy. To address this issue, we propose a cross-modal ultra-scale learning network (CMUS-Net) for the auxiliary diagnosis of multiple glomerular diseases. CMUS-Net utilizes multiple ultrastructural information to bridge the scale difference between nanometer and micrometer images. Specifically, we introduce a sparse multi-instance learning module to aggregate features from TEM images. Furthermore, we design a cross-modal scale attention module to facilitate feature interaction, enhancing pathological semantic information. Finally, multiple loss functions are combined, allowing the model to weigh the importance among different modalities and achieve precise classification of glomerular diseases. Our method follows the conventional process of renal biopsy pathology diagnosis and, for the first time, performs automatic glomerular multi-disease classification based on images from three modalities and two scales. On an in-house dataset, CMUS-Net achieves an ACC of 95.37$\pm$2.41\%, an AUC of 99.05$\pm$0.53\%, and an F1-score of 95.32$\pm$2.41\%. Extensive experiments demonstrate that CMUS-Net outperforms other well-known multi-modal or multi-scale methods and show its generalization capability in staging MN. Code is available at \url{https://github.com/SMU-GL-Group/MultiModal_lkx/tree/main}.
\end{abstract}

%% main text
\section{Introduction}
\label{section1}
\begin{figure*}[htb]
	\centering
	\setlength{\abovecaptionskip}{0.1cm}
	\includegraphics[width=\textwidth]{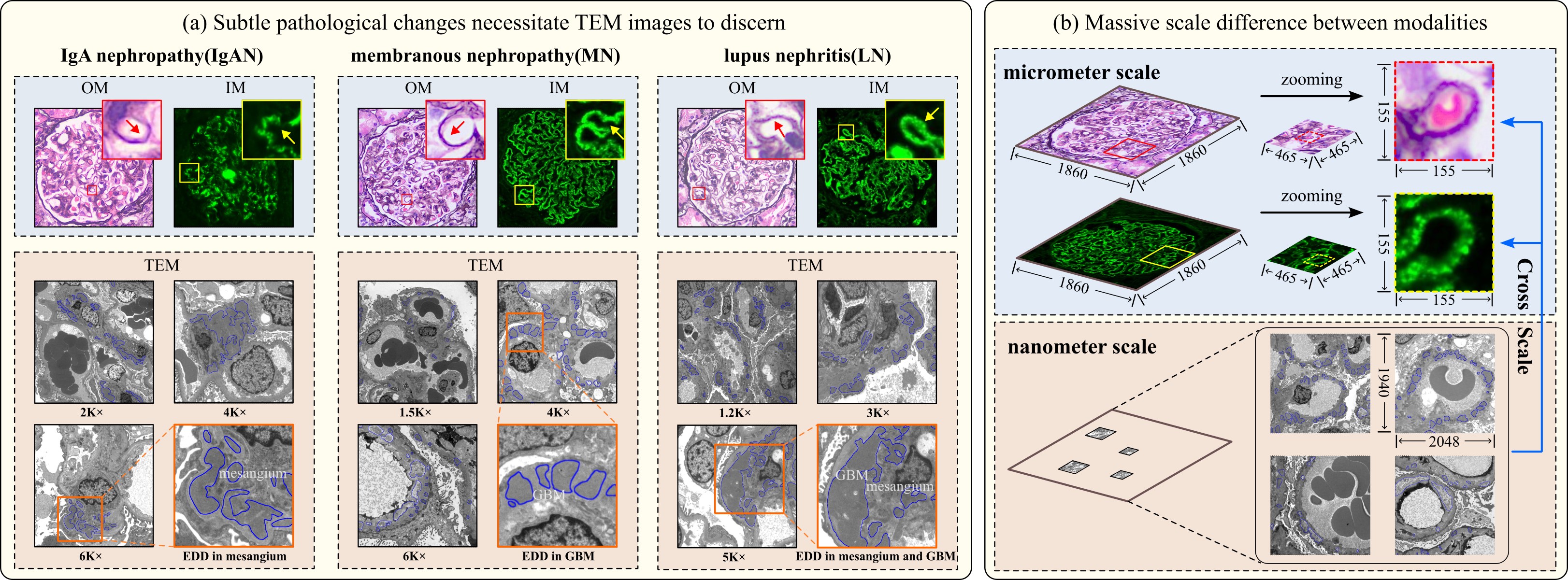}
	\caption{Illustration of major challenges in classifying glomerular diseases based on renal biopsy images: (a) Subtle pathological changes necessitate TEM images to discern. For example, the deposition locations of immune complexes vary among different diseases, with deposits occurring in the mesangium for IgAN, the glomerular basement membrane (GBM) for MN, and both locations for LN. These details are difficult to distinguish under OM and IM (as indicated by arrows), and precise identification of the deposition locations necessitates TEM. The blue lines in TEM images outline the main areas of these immune complexes deposition, also named electron-dense deposits (EDDs) under TEM. (b) Massive scale difference between modalities. After zooming the same type of detailed lesion areas (such as GBM) in OM and IM images to match those in TEM images, the resolution remains low, and it is still difficult to clearly identify the exact deposition locations of EDDs.}
	\label{fig1}
\end{figure*}

Glomerular diseases are the leading cause of chronic and end-stage renal failure \citep{marshall2006cell}. It is complex and diverse and can be categorized into various types based on its etiology, pathological characteristics, and clinical manifestations \citep{anders2023glomerulonephritis}. The accurate diagnosis of glomerular diseases relies on renal biopsy \citep{jennette1997diagnosis}. Renal biopsy is the gold standard for diagnosing glomerular diseases \citep{luciano2019update}, which includes three types of microscopic examinations: optical microscopy (OM), immunofluorescence microscopy (IM), and transmission electron microscopy (TEM) \citep{walker2004practice}. In the standard renal biopsy workflow, pathologists must repeatedly observe renal tissues with the naked eye and search for and identify subtle morphology changes, which is a time-consuming and labor-intensive process, limiting diagnostic efficiency. There is an urgent need to leverage deep learning methods to construct automated disease classification models based on images obtained from these three examinations to assist pathologists in efficient diagnosis. Current deep learning research primarily focuses on integrating OM and IM images to classify glomerular diseases. OM and IM images reveal morphological alterations and histochemical features of renal tissues at the micrometer scale. However, the histopathological manifestations of glomerular diseases are intricate, with some subtle lesions being difficult to identify accurately in OM or IM images, necessitating TEM images to discern \citep{angelotti2021imaging,najafian2022approach,pavlisko2013continued,pease1955fine,muehrcke1969clinical,mccluskey1971value} (Fig.~\ref{fig1} (a)). Therefore, it is essential to integrate TEM images with OM and IM images to construct a multi-modal model for accurately classifying various glomerular diseases \citep{huang2020review}.

Multi-modal learning can integrate the complementary information from various modalities and model their associations with diseases or lesions \citep{baltruvsaitis2018multimodal}, which has been applied successfully to the auxiliary diagnosis of breast cancer, skin lesions, Alzheimer's disease, and other diseases \citep{sun2023scoping,li2023artificial,cheerla2019deep,liu2024hammf}. However, unlike multi-modal information fusion between tissue-level images such as CT, MR, ultrasound, and whole slide images \citep{shi2021c,sedghi2020improving,zhang2022mmformer,xia2019novel}, three types of images from renal biopsy involve both tissue and cell levels, covering micrometer and nanometer scales. This demands the model to consider their scale differences while fusing information from different modalities, posing a challenge to existing multi-modal learning methods.  A direct response to this challenge is to combine multi-scale learning methods. Multi-scale learning methods can utilize receptive fields of different sizes to expand the perception range of the model, thereby capturing the features of images at different scales \citep{olimov2023consecutive,fu2021multiscale,qiu20243d}. However, these methods hinge on spatial correspondence within the image and are more sensitive to the scale differences in the same modality. They struggle to perceive the massive scale differences between various modalities. After zooming the same type of detailed lesion areas in OM and IM images to match those in TEM images, the resolution remains low, and it is still difficult to clearly identify the subtle pathological changes (Fig.~\ref{fig1} (b)). To effectively integrate features across these three modalities, it is essential to solve the problem of cross-modal scale feature fusion.

In this study, we propose a new multi-modal early-feature fusion method named cross-modal ultra-scale learning network (CMUS-Net) to address the above issues. The key innovation of this method lies in aligning information from multiple nanometer-scale images with the ones from micrometer-scale images and combining attention mechanisms to achieve cross-modal scale feature fusion between OM, IM, and TEM modalities.

Our study makes the following contributions:
\begin{itemize} 
	\item{} Few studies have attempted to construct multi-modal models by integrating ultramicroscopic images with microscopic images. We propose the CMUS-Net framework, which is based on TEM, OM, and IM modalities to achieve automated glomerular multi-disease classification.
    
	\item{} We introduced a sparse multi-instance learning module to achieve cross-modal scale alignment by aggregating information from multiple TEM images, thereby bridging the scale difference between modalities.
    
	\item{} We propose a cross-modal scale attention module that leverages aligned features to enhance relevant semantic information across modalities, thereby strengthening the representation of pathological features.
    
	\item{} Experiments on an in-house glomerular multi-disease classification dataset demonstrate that the proposed method outperforms other multi-modal multi-scale methods. Besides disease classification, our method can be applied to other analysis tasks of renal biopsy images, such as membranous nephropathy(MN) staging.
\end{itemize}

\section{Related works}
\label{section2}
\subsection{Deep learning based on renal biopsy images}
\label{section2-1}
The renal biopsy images from OM, IM, and TEM provide complementary features from the perspectives of tissue morphology, immunohistochemistry, and ultrastructural pathology, respectively, while containing relevant pathological semantic information. For example, typical pathological changes of MN include thicker glomerular capillary walls and spike-like projections appearing in OM images, immune complexes along the capillary wall appearing in IM images, and electron-dense deposits (EDDs) on the basement membrane presenting in TEM images \citep{lai2015membranous,xu2024fate}. They are the changes of the same pathological characteristics under different perspectives. Fully leveraging these complementary features and relevant pathological semantic information can facilitate the accurate classification of glomerular disease.

Currently, there are some auxiliary diagnosis studies based on renal biopsy images. Hao et al. \citep{hao2023accurate} combined OM and IM images to identify membranous nephropathy; Wang et al. \citep{wang2023ada} and Kitamura et al. \citep{kitamura2020deep} used IM images obtained from different antibody detection to classify membranous nephropathy and diabetes nephropathy respectively. Yasmine et al. \citep{abu2022development} classified renal cell carcinoma subtypes. Despite the achievements of the abovementioned methods, their limitations are evident: they only utilize micrometer-scale pathological information and target a single disease, which does not yet meet clinical needs. To assist glomerular disease diagnosis, a multi-disease classification model can be constructed, incorporating multi-modal learning methods to integrate the nanometer-scale TEM images with micrometer-scale OM and IM images.

\subsection{Disease auxiliary diagnosis based on multi-modal learning}
\label{section2-2}
Multi-modal learning can integrate images from different sources, effectively fuse complementary information between modalities, and assist pathologists in diagnostic analysis \citep{duan2024deep,dai2021transmed,zhou2023transformer}. Furthermore, although the imaging methods for different medical images vary, these images contain relevant semantic information. Combining multi-modal learning methods with attention mechanisms can effectively fuse complementary information and enhance relevant semantic information between modalities. For instance, Song et al. \citep{song2022cross} proposed a cross-modal attention mechanism that explicitly models the spatial correspondence between MR and TRUS images to guide the registration of prostate cancer biopsy images. Inspired by channel attention \citep{Hu2017SqueezeandExcitationN}, Jian et al. \citep{jian2021multiple} treated different MR modalities as distinct channels and constructed a modality-based attention module to classify borderline and malignant epithelial ovarian tumors. Golovanevsky et al. \citep{golovanevsky2022multimodal} used a bidirectional cross-modal attention mechanism to model the interaction between MR images and clinical and genetic data modalities.

The multi-modal images involved in the methods above are all derived from the patient's identical tissue regions and are at the same scale level. When applying attention mechanisms to model the feature interactions between these images, it is unnecessary to consider whether they are aligned in scale. However, the scale of renal biopsy images encompasses both micrometer and nanometer scales (Fig.~\ref{fig1}), and the three images come from different slices of renal tissue from the same patient, making it impossible to match them directly in space. Therefore, when performing multi-modal feature fusion on renal biopsy images, it is crucial to align the scale differences between OM, IM images, and TEM images in addition to fusing complementary information and enhancing relevant semantic information. To address this issue, one might consider incorporating multi-scale learning methods as a solution.

\begin{figure*}[htb]
	\centering
	\includegraphics[width=\textwidth]{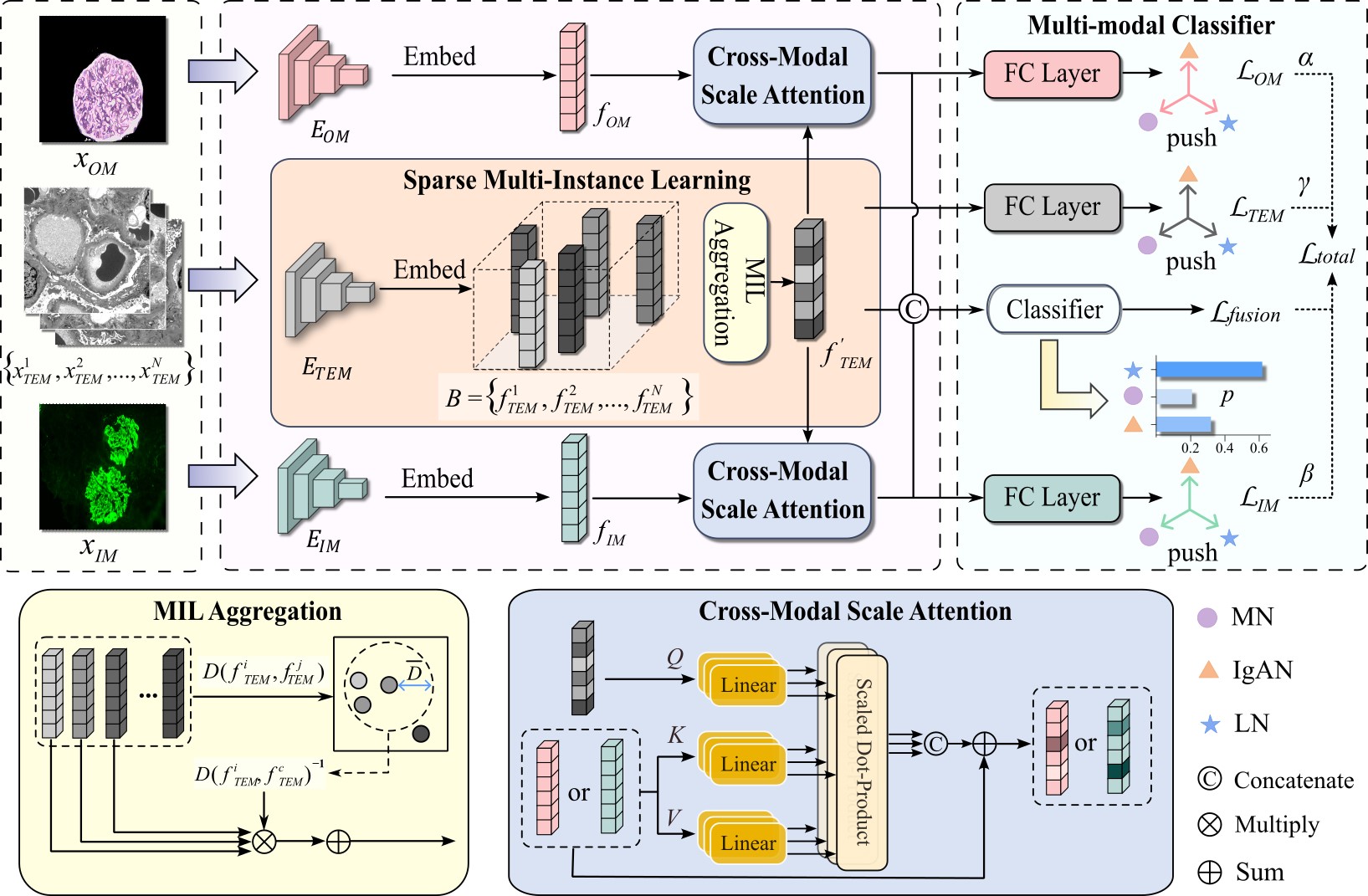}
	\caption{The overview of the proposed CMUS-Net framework for glomerular disease classification based on multi-modal renal biopsy images.}
	\label{fig2}
\end{figure*}

\subsection{Disease auxiliary diagnosis based on multi-scale learning}
\label{section2-3}
Multi-scale learning leverages the diverse scale features within medical images to boost the model's scale awareness \citep{gurcan2009histopathological} and has emerged as a research hotspot in disease auxiliary diagnosis \citep{diao2023deep}. For instance, some methods acquire multi-scale features at various magnification levels through image pyramids to aid in diagnosing diseases such as ascending colon disease, glioma, and breast cancer \citep{gao2022automatic,liu2024multi,Li2020DualstreamMI,deng2024cross}. Although these methods facilitate the interaction of multi-scale features, the scales involved are still within the micrometer range of magnifications from $5\times$  to $40\times$, and they only consider features from a single modality. Some research has combined multi-modal and multi-scale approaches to obtain complementary features across modalities and detailed features at distinct scales. For example, He et al. \citep{he2021multi} proposed the MSAN, with a feature pyramid structure, to capture features at different resolutions in retinal images and fuse them with those from OCT images. Wu et al. \citep{wu2023aggn} introduced the AGGN, which utilizes a multi-scale feature extraction module to obtain fine-grained texture and tissue information from glioblastoma MR images and integrates information from different sequence images.

The above multi-modal multi-scale methods focus on perceiving and capturing different scale features within a single modality and gradually merge features from large to small scale, allowing the model to pay attention to detailed information. However, if the scale differences between modalities are too huge, it will exceed the information perception range of these multi-scale methods. Renal biopsy images include OM and IM images at micrometer-scale magnifications ranging from $200\times$ to $400\times$, as well as TEM images at nanometer-scale magnifications ranging from $1K\times$ to tens of $K\times$ \citep{hacking2021deep}. It is challenging to capture the ultrastructural information in TEM images by applying multi-scale methods to OM and IM images (as shown in Fig.~\ref{fig1} (b)). Therefore, we propose the CMUS-Net framework to bridge the scale differences by aggregating information from multiple nanometer-scale images with the ones from micrometer-scale images to achieve cross-modal scale feature fusion.

\section{Method}
\label{section3}
Fig.~\ref{fig2} shows the framework of the proposed method. It addresses the primary challenges associated with the insufficiency of existing glomerular disease classification models in utilizing ultrastructural information of renal biopsy images, as well as feature fusion between nanometer- and micrometer-scale images.

\subsection{Problem formulation}
\label{section3-1}
Our study aims to enhance the model's performance in classifying glomerular diseases by effectively integrating OM, IM, and TEM modalities, thereby assisting pathologists in diagnosis. Let $X = \left\{(x_{OM}, x_{IM}, x_{TEM}, y)_{m}\right\}_{m=1}^M$ represent a dataset containing $M$ patients with three types of images, and $y$ represents the patient's disease category label. In general, a multi-modal model $P$ for the automatic classification of glomerular diseases can be formulated as

\begin{equation}\label{eq1}
p=P(x_{k})
\end{equation}

\noindent where $p$ is the model’s prediction probability, and $k \in$ [OM, IM, TEM] denotes the type of image utilized by the model. Given that there is a significant scale difference between nanometer-scale TEM images and micrometer-scale OM and IM images, integrating them poses a considerable challenge. Most existing multi-modal classification models based on renal biopsy images generally leverage OM and IM images only. Moreover, due to the absence of ultrastructural information on glomeruli provided by TEM images, these models mainly target a single disease. To address these challenges, we have developed a model with tri-modalities of renal biopsy images to classify multiple glomerular diseases, which is defined as

\begin{equation}\label{eq2}
p=P(CMSA(x_{OM},x_{IM},SMIL(x_{TEM}))
\end{equation}

According to Eq.~(\ref{eq2}), our proposed CMUS-Net framework mainly incorporates a sparse multi-instance learning (SMIL) module to bridge the scale differences between TEM images and OM and IM images, with a cross-modal scale attention (CMSA) module to enhance their feature interaction.

\subsection{CMUS-Net}
\label{section3-2}
The architecture of our proposed CMUS-Net framework is depicted in Fig.~\ref{fig2}, comprising three primary components: a SMIL module, a pair of CMSA modules, and a multi-modal classifier. Specifically, we introduce the SMIL module to aggregate features from multiple TEM images. It aligns the scale information from micrometer- and nanometer-scale images in the feature space. Subsequently, two CMSA modules facilitate feature interaction between the features of the TEM modal and those of the OM or IM modal, respectively. This interaction strengthens the relevant pathological semantic information. Finally, the multi-modal classifier receives features from each modality as well as the fused one to identify various glomerular diseases. Below, we will provide a detailed introduction to each component of the proposed framework.

\subsubsection{SMIL module}
\label{section3-2-1}
We employ three encoders—$\textbf{E}_{OM}$, $\textbf{E}_{IM}$, and $\textbf{E}_{TEM}$—to extract features from the patient's OM image $x_{OM}$, IM image $x_{IM}$, and TEM image $x_{TEM}$, respectively. A major challenge in achieving feature fusion of OM, IM, and TEM images is the significant scale difference between them. TEM images offer exceptionally high magnification, providing detailed structural information of the glomerulus at the nanometer scale. However, this high magnification comes with a limited field of view (FOV). To comprehensively assess a patient's lesions, pathologists must integrate information from multiple FOVs to make accurate judgments. Therefore, when observing through TEM, pathologists identify and document typical lesion regions across various FOVs, saving these as multiple images $\left\{x_{TEM}^1,x_{TEM}^2,...,x_{TEM}^n\right\}$, where $n$ represents the number of patient’s TEM images. We utilize the SMIL module to aggregate the features of these TEM images, thereby capturing comprehensive pathological information for each patient.

MIL is commonly applied to analyze whole slide images (WSI), where it categorizes lesion patches as positive instances and normal tissue patches as negative instances. All these instances within a single WSI constitute a bag. The model analyzes the images in a weakly supervised manner by identifying positive instances within each bag \citep{carbonneau2018multiple}. Unlike the tens of hundreds of patches in a WSI \citep{shao2021transmil}, the number of TEM images acquired by pathologists is typically only a few. These TEM images predominantly contain lesion regions and are thereby classified as positive instances. Consequently, we introduce the Sparse Multi-Instance Learning (SMIL) module, as detailed in Fig.~\ref{fig2} which treats the feature of an individual TEM image as an instance, denoted as $f_{TEM}^{j}$, $j \in (1,n]$ and all instances from the same patient as a bag, denoted as

\begin{equation}
\label{eq3}
\begin{aligned}
B &=\left\{f_{TEM}^{1}, f_{TEM}^{2}, ..., f_{TEM}^{n}\right\} \\
  &=\textbf{E}_{TEM}(x_{TEM}^1,x_{TEM}^2,...,x_{TEM}^{n})
\end{aligned}
\end{equation}

The SMIL module leverages only a limited number of instances to aggregate critical pathological information and yield the TEM modal features. First, as shown in Eq.~(\ref{eq4})-(\ref{eq5}), we calculate the Euclidean distances $D(f_{TEM}^{i},f_{TEM}^{j})$ between instances to identify the central instance $f_{TEM}^{c}$, which has the minimum sum of distances to all other instances, and compute the mean distance $\overline{D}$ among all instances. Subsequently, to mitigate the impact of outliers on the overall TEM modal features, we exclude those instances whose distances from the central instance exceed a certain threshold  $t\overline{D}$. Following this, the remaining instances, excluding the central one, are weighted based on their distances to the central instances. Finally, the weighted instances are aggregated with the central one, as illustrated in Eq.~(\ref{eq6})-(\ref{eq7}):

\begin{equation}\label{eq4}
D(f_{TEM}^{i},f_{TEM}^{j})=\sqrt{\sum_{k=1}^{d}(f_{TEM}^{i,k},f_{TEM}^{j,k})^{2}}
\end{equation}

\begin{equation}\label{eq5}
\overline{D}=\frac{2}{n(n-1)} \sum_{i=1}^{n-1} \sum_{j=i+1}^{n} D(f_{TEM}^{i},f_{TEM}^{j})
\end{equation}

\begin{equation}\label{eq6}
B^{\prime}=\left\{f_{TEM}^{c},f_{TEM}^{i}|D(f_{TEM}^{c},f_{TEM}^{i}) \leq t\overline{D},i \neq c\right\}
\end{equation}

\begin{equation}\label{eq7}
f_{TEM}^{\prime}=f_{TEM}^{c}+\sum_{i=1,i \neq c}^{n}(w_{i} f_{TEM}^{i}), f_{TEM}^{i} \in B^{\prime}
\end{equation}

\noindent where $f_{TEM}^{\prime}$ represents the TEM modal features, $t$ is set to $1.5$ in the experiment. $w_{i}$ represents the normalized distance weight, with the distance weights set as the inverse of the distances from the central instance $D(f_{TEM}^{c}, f_{TEM}^{i})$. By leveraging the SMIL module to aggregate features from nanoscale images, our classification model bridges the scale differences between TEM and OM, IM images in the feature space, enabling smooth feature fusion across modalities.

\subsubsection{CMSA module}
\label{section3-2-2}
As depicted in Fig.~\ref{fig1}, despite the lack of direct spatial correspondence between the OM, IM, and TEM modalities, each image provides complementary features from distinct perspectives: tissue morphology, immunohistochemistry, and ultrastructural pathology, respectively. In addition, these modalities offer relevant pathological semantic features at both micrometer and nanometer scales. The multi-modal model needs to enhance these relevant features while integrating the complementary ones. Therefore, we utilize the ultrastructural pathological features provided by multiple TEM images at the nanometer scale to identify tissue morphology or immunohistochemical features reflecting relevant lesions at the micrometer scale. Finally, we integrate the output to facilitate the feature interaction process.

The feature interaction process is detailed Cross-Modal Scale Attention (CMSA) module of Fig.~\ref{fig2}. The extracted features of each modality are input into the CMSA module. We then leverage the TEM modal features $f_{TEM}^{\prime}$ to calculate attention scores with OM modal features $f_{OM}$ and IM modal features $f_{IM}$, respectively. This approach aims to enhance the relevant pathological semantic features across different modalities. To elucidate this process, we take the interaction between OM and TEM modalities as an example. Initially, the TEM modal features $f_{TEM}^{\prime}$ are encoded as query vectors $Q_{TEM}\in R^d$, while the OM modal features $f_{OM}$ are encoded as both key vectors $K_{OM}\in R^d$ and value vectors $V_{OM}\in R^d$. Subsequently, we obtain the weighted OM modal features by applying multi-head attention, as detailed in Eq.~(\ref{eq8})-(\ref{eq9}):

\begin{equation}
    \label{eq8}
head_{OM}^{h}=Softmax(\frac{Q_{TEM}^{h}(K_{OM}^{h})^{T}}{\sqrt{d/h}})V_{OM}^{h}
\end{equation}

\begin{equation}
	\label{eq9}
f_{OM}^{\prime}=Concat(head_{OM}^{1},...,head_{OM}^{h})+f_{OM}
\end{equation}

\noindent 
where $d$ represents the dimension of the vectors, and $h$ denotes the number of attention heads which is set to 4 in our experiment. Similarly, we can obtain the weighted features $f_{IM}^{\prime}$ for the IM modal:

\begin{equation}
    \label{eq10}
f_{IM}^{\prime} = Concat(head_{IM}^{1},...,head_{IM}^{h})+f_{IM}
\end{equation}

\subsubsection{Multi-modal classification}
\label{section3-2-3}
After concatenating the TEM modal features $f_{TEM}^{\prime}$, the weighted OM modal features $f_{OM}^{\prime}$, and the weighted IM modal features $f_{IM}^{\prime}$ to yield the fused features $f_{fusion}$, a Classifier consisting of two fully connected layers is employed to obtain the final prediction probability $p$, as shown in Eq.~(\ref{eq11}) and~(\ref{eq12}):

\begin{equation}
    \label{eq11}
f_{fusion} = Concat\left( f_{OM}^{\prime}, f_{TEM}^{\prime}, f_{IM}^{\prime}\right)
\end{equation}

\begin{equation}
    \label{eq12}
p = Classifier(f_{fusion})
\end{equation}

In our CMUS-Net framework, the encoders $\textbf{E}$ and the Classifier are interchangeable. This means that they can be updated accordingly when other SOTA classification models become available. As a result, our framework can continuously benefit from advancements in classification model performance, thereby enhancing its flexibility and adaptability.

\subsection{Weighted Loss Function}
\label{section3-3}
The cross-entropy loss is employed in our CMUS-Net, as detailed in Eq.~(\ref{eq13}).

\begin{equation}
    \label{eq13}
\mathcal L_{fusion} = -y\log(p)-(1-y)\log(1-p)
\end{equation}

\noindent where $y$ represents the ground-truth label.

In the diagnosis and analysis of glomerular diseases, information from each modality may contribute differently. To enable the model to perceive the relative importance of different modalities, we adopt a simplified and practical approach, as proposed by \cite{he2021multi}, to train our multi-modal model. Specifically, in addition to the loss function $\mathcal L_{fusion}$, we introduce separate fully connected layers $\textbf{FC}$ for features $f_{OM}^\prime$, $f_{IM}^\prime$, and $f_{TEM}^\prime$ to obtain their respective loss functions, as shown in Eq.~(\ref{eq14}). Therefore, the total loss $\mathcal L_{total}$ is presented in Eq.~(\ref{eq15}).

\begin{equation}
\label{eq14}
\mathcal L_{k} = -y\log(\textbf{FC}(f_{k}^\prime))-(1-y)\log(1-\textbf{FC}(f_{k}^\prime))
\end{equation}

\begin{equation}
\label{eq15}
\mathcal L_{total} = \alpha \mathcal L_{OM}+\beta \mathcal L_{IM}+\gamma \mathcal L_{TEM}+\mathcal L_{fusion}
\end{equation}

\noindent 
where $k \in$ [OM, IM, TEM]. $\alpha$, $\beta$, and $\gamma$ are weighting coefficients such that $\alpha+\beta+\gamma=1$. These coefficients determine the relative importance of each modality in the total loss function.

\section{Experiments}
\label{section4}
\subsection{Data}
\label{section4-1}
This study was a retrospective investigation approved by the Institutional Review Board, exempt from informed consent with the guiding principles of the Declaration of Helsinki. All patient data were anonymized before analysis. We collected renal biopsy data from the Guangzhou Huayin Medical Laboratory Center between December 2021 and December 2023, which included three common glomerular diseases: IgA nephropathy (IgAN), membranous nephropathy (MN), and lupus nephritis (LN), with respective incidence rates of 28.1\%, 23.4\%, and 13.5\% \citep{xu2016long}.

Applying staining methods with hematoxylin-eosin (H\&E), periodic acid-Schiff (PAS), periodic acid-silver methenamine (PASM), and other staining protocols can obtain different OM images. Given that identical tissue structures exhibit varying colors under different staining protocols, to prevent potential impacts on the model performance, we exclusively utilized OM images with PASM for our experiments. Additionally, since the critical structural lesions characteristic of glomerular diseases are predominantly evident in the glomerular region, we employ SAM \citep{Kirillov2023SegmentA} to conduct semi-automatic segmentation of the glomerular region in all OM images. Immunofluorescence staining with antibodies, including immunoglobulin A (IgA), immunoglobulin G (IgG), and complement C3, produces various IM images. Notably, IM images that show positive antibody detection have higher immunofluorescence intensity, enabling the model to learn more about the patterns of immune-complex deposits from these images. Consequently, for the experiments in this study, we exclusively utilized IM images with positive antibody detection results. To observe the ultrastructural changes in IgAN, MN, and LN, pathologists use TEM examinations at magnifications of 1$K\times$ to 27$K\times$ to identify typical lesion areas, resulting in several TEM images covering different tissue regions. Therefore, we included all these TEM images for each patient.

\begin{table}[h]
	\caption{\label{table1}Dataset statistical information. Parentheses indicate the types of staining protocols/antibodies used for the images.}
	\centering
	\small
	%\resizebox{\linewidth}{!}{
		\begin{threeparttable}
			%\begin{tabular}{lllll}
			\begin{tabularx}{\linewidth}{@{}XXlll@{}}
				\toprule
				\multirow{2}{*}{Datasets} & \multirow{2}{*}{Categories} & \multicolumn{3}{c}{Modalities} \\
				                                                      \cmidrule{3-5}
				     &                                                & OM        & IM        & TEM  \\
				\midrule
				\multirow{3}{*}{Diseases}
				     & IgAN                                           & 182(PASM) & 182(IgA)  & 1568 \\
				     & MN                                             & 177(PASM) & 177(IgG)  & 1518 \\
				     & LN                                             & 182(PASM) & 182(IgG)  & 1659 \\
				\midrule
				\multirow{3}{*}{Stages}
				     & MN-\uppercase\expandafter{\romannumeral1}      & 52(PASM)  & 52(IgG)   & 413  \\
				     & MN-\uppercase\expandafter{\romannumeral2}      & 143(PASM) & 143(IgG)  & 1518 \\
				     & MN-\uppercase\expandafter{\romannumeral3}      & 80(PASM)  & 80(IgG)   & 1659 \\
				\bottomrule
			\end{tabularx}
		\end{threeparttable}
	%}
\end{table}

\begin{table*}[htb]
	\caption{\label{table2}Ablation experiment of the SMIL module, CMSA module and the WL function. \textbf{Bold} denotes the best result in each column. The \textit{\textbf{p}}-value displays the paired t-test results compared to the AUC value of the reference (\textbf{Ref.}).}
	\centering
	\small
	\resizebox{\linewidth}{!}{
		\begin{tabular}{@{}l l *{8}{c}@{}}
			\toprule 
			SMIL       & CMSA       & WL         & ACC(\%)        & AUC(\%)        & PRE(\%)        & REC(\%)        &SPE(\%)         & F1-score(\%)   & $p$-value \\
			\midrule 
			           &            &            & 91.30$\pm$1.72 & 97.77$\pm$0.69 & 91.33$\pm$1.27 & 91.59$\pm$1.28 & 95.80$\pm$2.89 & 91.21$\pm$1.50 & 0.000 \\
			           &            & \checkmark & 92.04$\pm$1.39 & 97.91$\pm$0.85 & 92.01$\pm$1.27 & 92.38$\pm$1.06 & 96.19$\pm$2.12 & 92.03$\pm$1.27 & 0.017 \\
			           & \checkmark &            & 92.22$\pm$2.31 & 97.85$\pm$0.59 & 92.24$\pm$1.95 & 92.51$\pm$1.77 & 92.96$\pm$1.67 & 92.18$\pm$2.05 & 0.004 \\
			\checkmark &            &            & 92.59$\pm$1.55 & 98.18$\pm$0.76 & 92.60$\pm$1.48 & 92.73$\pm$1.72 & 96.36$\pm$1.50 & 92.53$\pm$1.56 & 0.029 \\
		  	           & \checkmark & \checkmark & 92.77$\pm$2.14 & 98.32$\pm$0.48 & 89.81$\pm$7.12 & 93.01$\pm$1.89 & 95.27$\pm$2.24 & 91.15$\pm$4.58 & 0.030 \\
			\checkmark &            & \checkmark & 93.15$\pm$2.39 & 98.73$\pm$0.30 & 93.16$\pm$2.26 & 93.53$\pm$2.21 & 96.76$\pm$2.52 & 93.17$\pm$2.33 & 0.163 \\
			\checkmark & \checkmark &            & 94.07$\pm$3.49 & 98.70$\pm$0.69 & 94.21$\pm$3.36 & 94.31$\pm$3.41 & 97.15$\pm$2.52 & 94.09$\pm$3.54 & 0.034 \\
			\checkmark & \checkmark & \checkmark & \textbf{95.37$\pm$2.41} & \textbf{99.05$\pm$0.53} & \textbf{95.29$\pm$2.46} & \textbf{95.52$\pm$2.29} & \textbf{97.78$\pm$1.67} & \textbf{95.32$\pm$2.41} & \textbf{Ref.} \\
			\bottomrule
		\end{tabular}
	%}
}
\end{table*}

\begin{figure*}[htb]
	\centering
	\includegraphics[width=\textwidth]{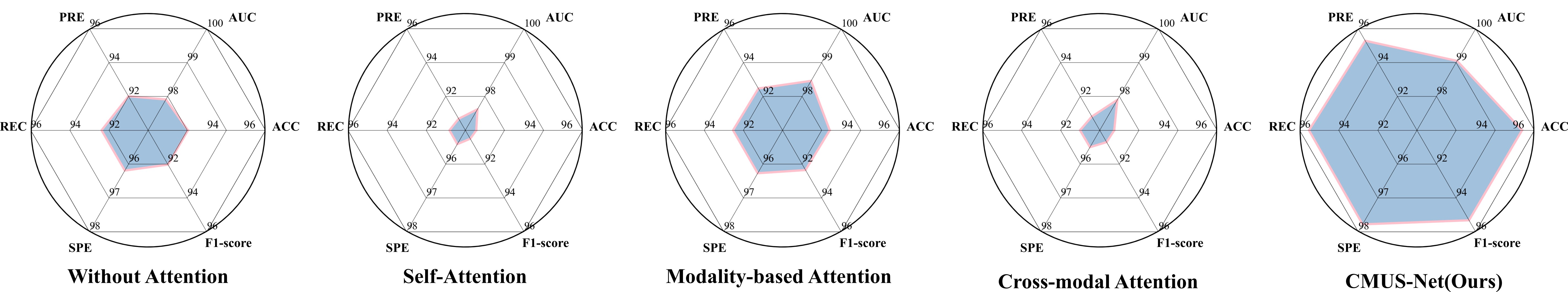}
	\caption{Comparison with other attention mechanisms.}
	\label{fig3}
\end{figure*}

Based on the outlined principles, we screened and constructed two in-house datasets from the collected data, with the number of images detailed in Table~\ref{table1}. Ultimately, each patient's data included one OM image, one IM image, and an average of 8 to 9 TEM images. Both the OM and IM images have a resolution of $3072\times2048$ pixels, with magnifications ranging from $200\times$ to $400\times$. The original sizes of the TEM image are $2048\times2048$ pixels. Following removing the scale bar and legend areas, the TEM images exhibit a resolution of $2048\times1940$ pixels with magnifications between 1$K\times$ to 27$K\times$.

\textbf{Multi-disease classification dataset}: The dataset includes patient data for individuals diagnosed with IgAN, MN, or LN. The disease category labels for the patients are based on the diagnostic results provided in their pathology reports. This dataset is utilized to assess the model performances of the glomerular multi-disease classification.

\textbf{MN staging dataset}: Besides multi-disease classification, we employed a staging task for MN to further assess our model’s performance. MN can be categorized into four stages according to its pathological progression, where each stage exhibits relatively distinct pathological characteristics and has well-defined diagnostic \citep{ronco2021membranous}. Since stage \uppercase\expandafter{\romannumeral4} of MN corresponds to the end stage of the disease and is associated with a very limited amount of biopsy data, we excluded the stage \uppercase\expandafter{\romannumeral4} MN data from our evaluation.

\subsection{Experimental settings}
\label{section4-2}
All experiments are performed with five-fold cross-validation, splitting each dataset into training and test sets in a 4:1 ratio. Each encoder of our model uses the SwinT-B \citep{Liu2021SwinTH} with ImageNet-1K pre-trained weights. Data augmentation techniques employed include random horizontal and vertical flipping. Input images are standardized, normalized, and resized to a resolution of $224\times 224$ to save training time and memory consumption. We use the Adam optimizer and update the learning rate with the cosine annealing strategy \citep{kirkpatrick1983optimization}. The initial learning rate and the weight decay factor are both set to $5\times 10^{-5}$. We train the models for 60 epochs to ensure convergence in all cases. Unless otherwise specified, the batch size is set to 4. Model performance is evaluated on test sets with accuracy (ACC), area under the receiver operating characteristic curve (AUC), precision (PRE), recall (REC), specificity (SPE), and F1-score. For an overall comparison of three diseases, we present the macro average of the metrics. All models are trained on an NVIDIA GeForce RTX 3090.

\subsection{Methods comparison}
\label{section4-3}
We compared our model with other multi-modal models, well-known unimodal classification models, and unimodal multi-scale models. First,we compare with several multi-modal models, including MMGL \citep{zheng2022multi}, mmFormer \citep{zhang2022mmformer}, MDL-IIA \citep{zhang2023predicting}, MSAN \citep{he2021multi}, and AGGN \citep{wu2023aggn}. These models originally receive different types of multi-modal inputs. (1) MMGL takes images, biomarkers, cognitive tests, and demographic data as inputs, extracting their features via transform matrices; (2) mmFormer utilizes the hybrid modality-specific encoders to extract features from four types of MRI images for brain tumor segmentation; (3) MDL-IIA accepts mammography images in both mediolateral oblique and craniocaudal views, alongside ultrasound images; (4) MSAN employs the multi-scale attention subnet and region-guided attention subnet to obtain features from fundus and OCT images, respectively; (5) AGGN receives four types of MRI images and leverages an equal number of convolutional branches to extract multi-scale features, followed by multi-stage feature fusion across modalities. To adapt these models to glomerular multi-disease classification, their inputs are replaced with OM, IM, and TEM images. Specifically, for mmFormer, we replace its segmentation head with two fully connected layers for classification. For MSAN, we employ its multi-scale attention subnet to extract features from each modality instead of the region-guided attention subnet, which is designed for 3D OCT images.

\section{Results}
\label{section5}
\subsection{Ablation studies}
\label{section5-1}
\subsubsection{Effectiveness of the modules and loss}
\label{Effectiveness of the modules and loss}
We conduct experiments on the multi-disease classification dataset to validate the effectiveness of the SMIL module, CMSA module, and the weighted loss function (WL) in our model. The results are presented in Table~\ref{table2}. When solely applying the Weighted Loss function, the model’s mean ACC, mean AUC, and mean F1-score improved by 0.74\%, 0.14\%, and 0.82\%, respectively. After incorporating the SMIL and CMSA modules, the model exhibits more pronounced performance improvement, with mean ACC, mean AUC, and mean F1-score increasing by 2.77\%, 0.93\%, and 2.88\%, respectively.

\subsubsection{Comparison with other attention mechanisms}
\label{section5-1-2}
To evaluate the capability of the SMIL and CMSA modules in promoting the feature interaction process, we compare them with other attention mechanisms commonly employed in multi-modal methods. Utilizing the multi-disease classification dataset, we train four models that simultaneously take OM, IM, and TEM modalities as inputs.  These models are based on our model but replace the SMIL and CMSA modules with (1) Without Attention, (2) Self-Attention \citep{Vaswani2017AttentionIA}, (3) Modality-based Attention \citep{jian2021multiple}, and (4) Cross-modal Attention \citep{golovanevsky2022multimodal}, respectively. Their performances are summarized in Fig.~\ref{fig3}. Our proposed method has achieved the best results across all metrics, attributed to the combination of SMIL and CMSA modules being able to explicitly guide the interaction process of features through the scale relationship between modalities.

\begin{figure*}[t]
	\centering
	\includegraphics[width=0.9\textwidth]{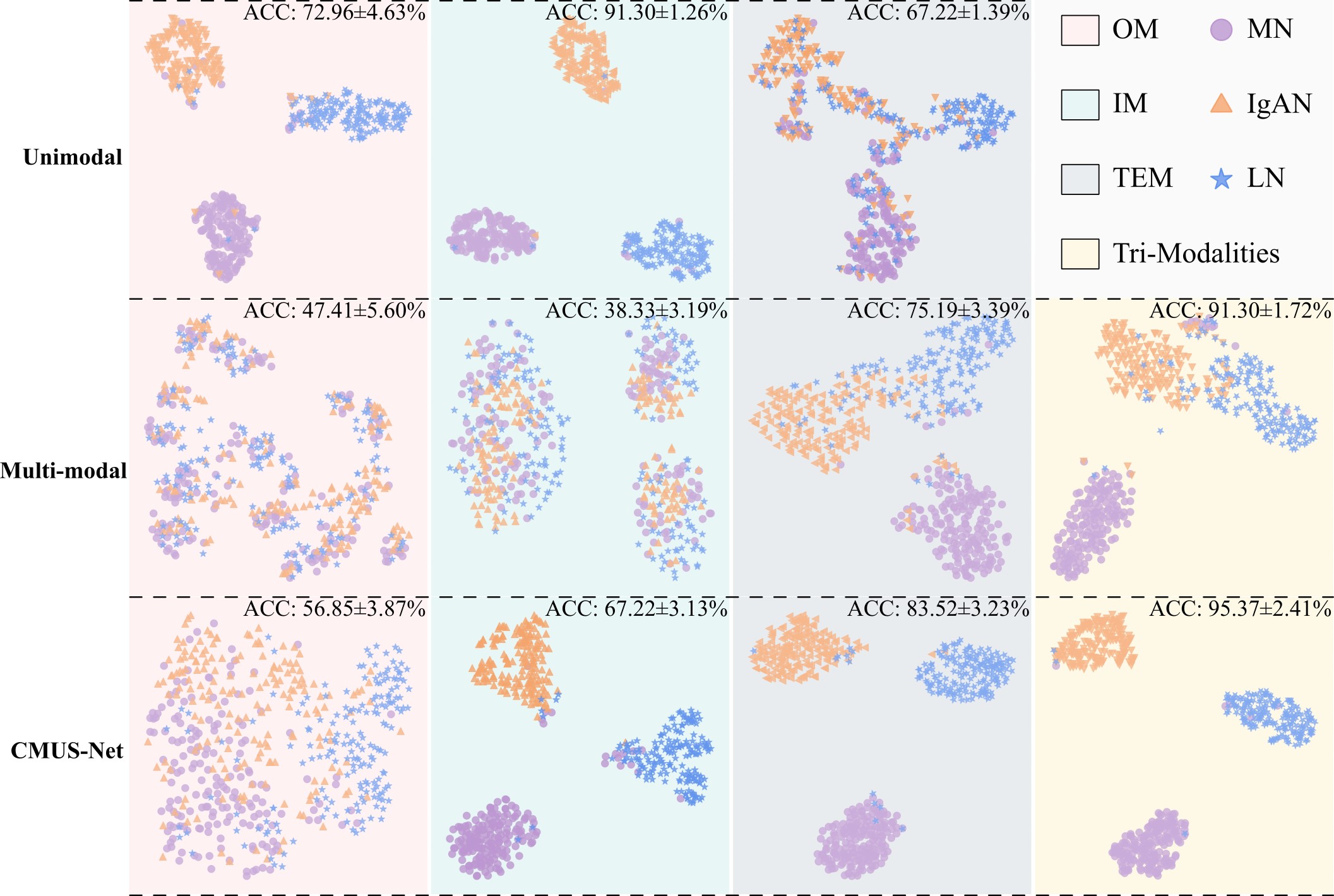}
	\caption{The t-SNE feature clustering results of encoders from different models. The rightmost column shows the clustering results of the fused features. ACC represents the classification accuracy.}
	\label{fig4}
\end{figure*}

\subsubsection{Clustering results of encoders}
\label{section5-1-3}
To fully illustrate the advantages of our proposed method in multi-modal feature interaction, we evaluate the feature extraction capabilities of encoders in unimodal and multi-modal models with the disease classification dataset. Initially, we construct a unimodal model by integrating a SwinT-B encoder with a two-layer multi-layer perceptron (MLP), which is trained with OM, IM, and TEM modalities as input. Following this, we apply t-SNE clustering to the features extracted by these trained encoders. The clustering results and classification accuracies, presented in the first row of Fig.~\ref{fig4}, provide both qualitative and quantitative insights into the encoders’ capability to extract and distinguish features for the three diseases. Subsequently, we compare it with a multi-modal model that also employs SwinT-B as the encoder for each modality. In this multi-modal model, the encoders extract features from each modality, which are then concatenated and fed into a two-layer MLP for classification. The clustering results of the features extracted by these modality encoders, as well as the classification accuracies after being fed into the MLP, are shown in the second row of Fig.~\ref{fig4}. Similarly, we obtain the clustering results of the modality encoders of CMUS-Net and their corresponding classification accuracies, as presented in the last row of Fig.~\ref{fig4}. It can be observed that compared to the mediocre multi-modal feature fusing method, CMUS-Net is able to effectively fuse cross-modal scale features, achieving an ACC of 95.37$\pm$2.41\%, which outperforms the best unimodal model and multi-modal model.

\begin{figure}[h]
	\centering
	\includegraphics[width=\columnwidth]{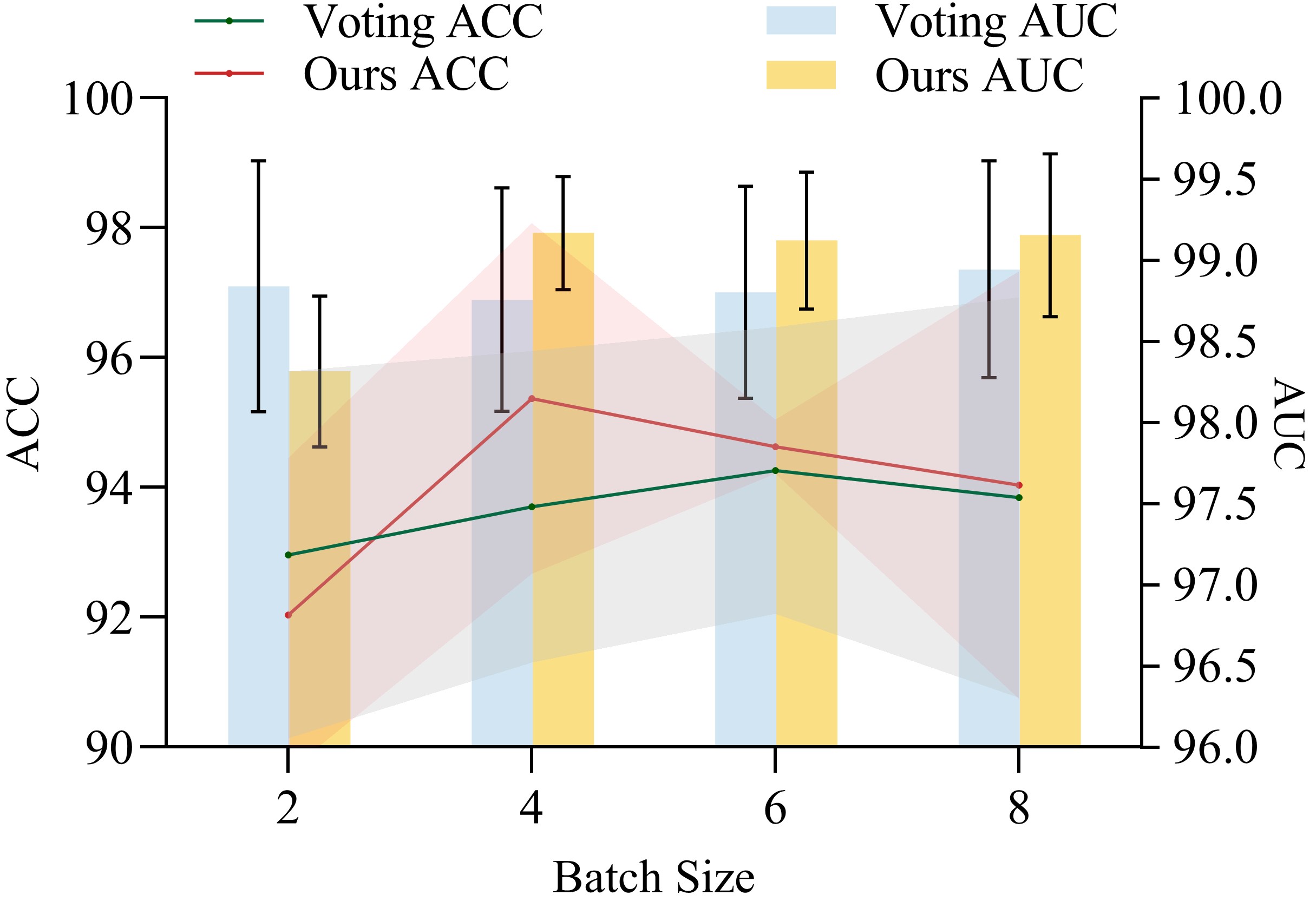}
	\caption{Comparison of our model with the voting-based late fusion model under different batch sizes. The shaded area of the polyline and the error line at the top of the bar represent the standard deviation range of ACC and AUC, respectively.}
	\label{fig5}
\end{figure}

\subsubsection{Comparison with late fusion method}
\label{section5-1-4}
Our proposed method integrates information from various modalities at the feature level, employing an early fusion approach. To demonstrate the superiority of early fusion over late fusion methods, we train a classification model, which employs a majority voting scheme to aggregate the prediction outcomes from OM, IM, and TEM modalities. All other configurations remain identical to those of our model. Fig.~\ref{fig5} illustrates that our model generally outperforms the voting-based late fusion model. This underscores the advantages of early fusion in enhancing the interaction of information from multiple modalities.

\begin{table*}[htb]
	\caption{\label{table3}Comparison results of different modal combinations. \textbf{Bold} denotes the best result in each column. The \textit{\textbf{p}}-value displays the paired t-test results compared to the AUC value of the reference (\textbf{Ref.}).}
	\centering
	\small
	\resizebox{\linewidth}{!}{
	\begin{tabular}{@{}l l *{8}{c}@{}}
			\toprule 
			OM           & IM         & TEM            & ACC(\%)        & AUC(\%)        & PRE(\%)        & REC(\%)        & SPE(\%)        & F1-score(\%)   & $p$-value \\
			\midrule 
			\checkmark &            &                & 72.96$\pm$4.63 & 87.99$\pm$3.03 & 72.67$\pm$5.10 & 72.47$\pm$5.30 & 86.28$\pm$4.15 & 72.39$\pm$5.28 & 0.000 \\
			           & \checkmark &                & 91.30$\pm$1.89 & 97.70$\pm$0.57 & 91.07$\pm$1.08 & 91.40$\pm$0.98 & 95.74$\pm$1.94 & 91.14$\pm$1.02 & 0.009 \\
			           &            & \checkmark     & 70.00$\pm$4.29 & 85.55$\pm$2.35 & 70.47$\pm$4.15 & 70.27$\pm$4.20 & 85.08$\pm$5.40 & 69.93$\pm$4.26 & 0.000 \\
			\checkmark & \checkmark &                & 92.04$\pm$1.72 & 97.93$\pm$0.58 & 92.10$\pm$1.57 & 92.19$\pm$1.65 & 93.89$\pm$1.12 & 91.96$\pm$1.66 & 0.000 \\
			\checkmark &            & \checkmark     & 91.11$\pm$3.64 & 97.78$\pm$0.64 & 91.04$\pm$3.69 & 91.31$\pm$3.54 & 92.01$\pm$1.98 & 91.00$\pm$3.64 & 0.001 \\
			           & \checkmark & \checkmark     & 94.26$\pm$1.08 & 98.19$\pm$0.71 & 94.16$\pm$0.98 & 94.41$\pm$0.90 & 93.79$\pm$1.45 & 94.19$\pm$0.96 & 0.041 \\
			\checkmark & \checkmark & \checkmark     & \textbf{95.37$\pm$2.41} & \textbf{99.05$\pm$0.53} & \textbf{95.29$\pm$2.46} & \textbf{95.52$\pm$2.29} & \textbf{97.78$\pm$1.67} & \textbf{95.32$\pm$2.41} & \textbf{Ref.} \\
			\bottomrule
		\end{tabular}
	%}
}
\end{table*}

\subsubsection{Comparison between different modal combinations}
\label{section5-1-5}
To analyze the impact of different modality combinations as input on model performance, we conduct ablation experiments on the multi-disease classification dataset. When input with OM, IM, or TEM modality, our model degenerates into a unimodal classification model. When input with OM and IM modalities, the model directly concatenates the extracted features for classification. The experimental results, shown in Table~\ref{table3}, indicate that the model achieves optimal performance when using OM, IM, and TEM modalities for inputs. Compared with the second best result, the mean ACC, mean AUC, and mean F1-score increased by 1.11\%, 0.86\%, and 1.13\%, respectively. This outcome underscores the indispensability of all three modalities for achieving precise auxiliary diagnosis of glomerular diseases.

\begin{figure}[htb]
	\centering
	\includegraphics[width=\columnwidth]{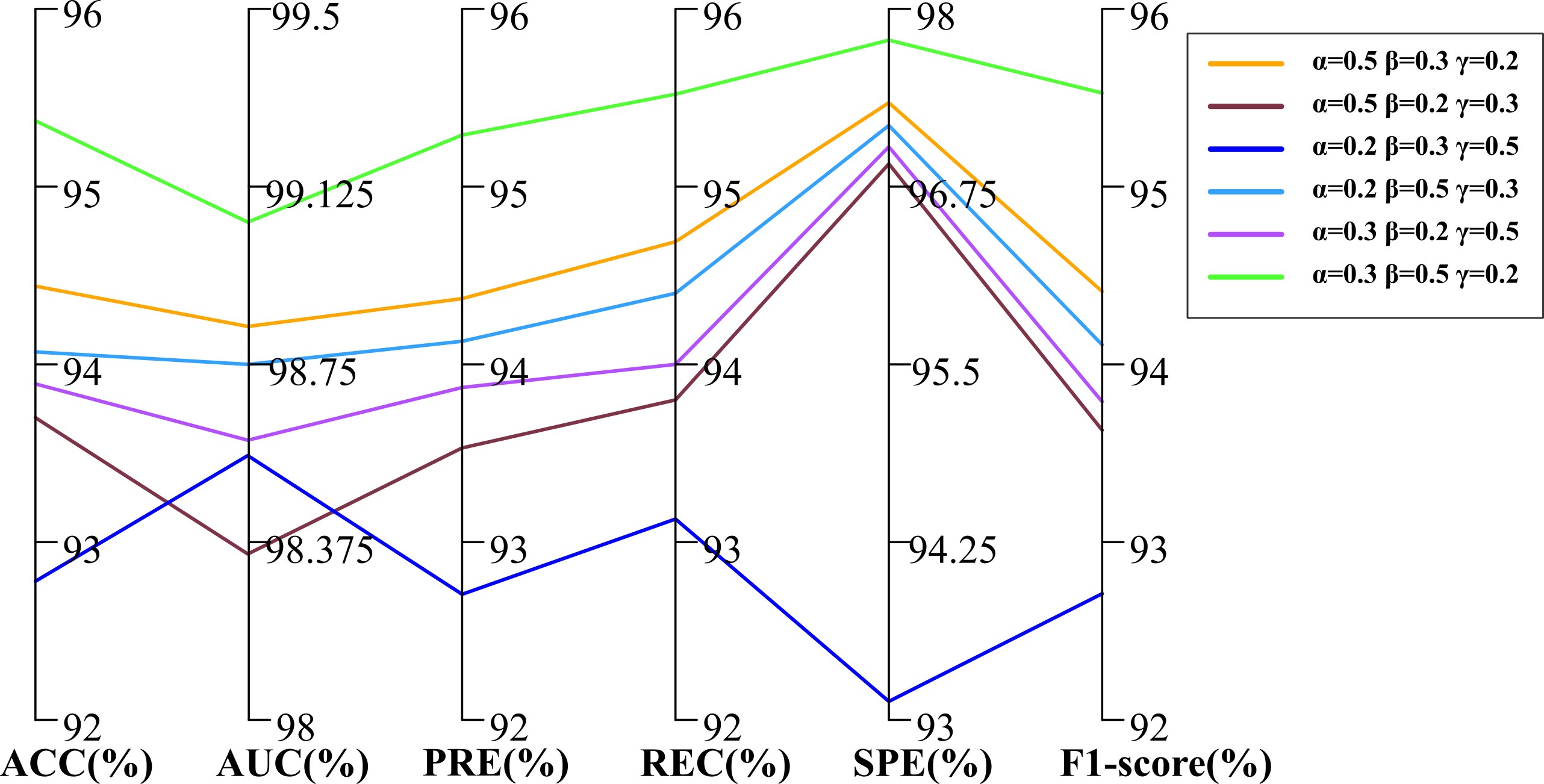}
	\caption{The performance of our model under different weighting coefficient settings.}
	\label{fig6}
\end{figure}

\subsubsection{Influence of weighting coefficients}
\label{section5-1-6}
Fig.~\ref{fig6} illustrates the performance variations of our model under different settings of weighting coefficients $\alpha$, $\beta$, and $\gamma$ in Eq.~(\ref{eq15}). Optimal performance is achieved with the weighting coefficients set to 0.5 for IM, 0.3 for OM, and 0.2 for TEM modalities. This also reflects the relative contribution of these three modalities in glomerular multi-disease classification to some extent.

\subsection{Comparison results on disease classification dataset}
\label{section5-2}
\subsubsection{Comparison with other models}
\label{section5-2-1}
\begin{figure*}[htb]
	\centering
	\includegraphics[width=\textwidth]{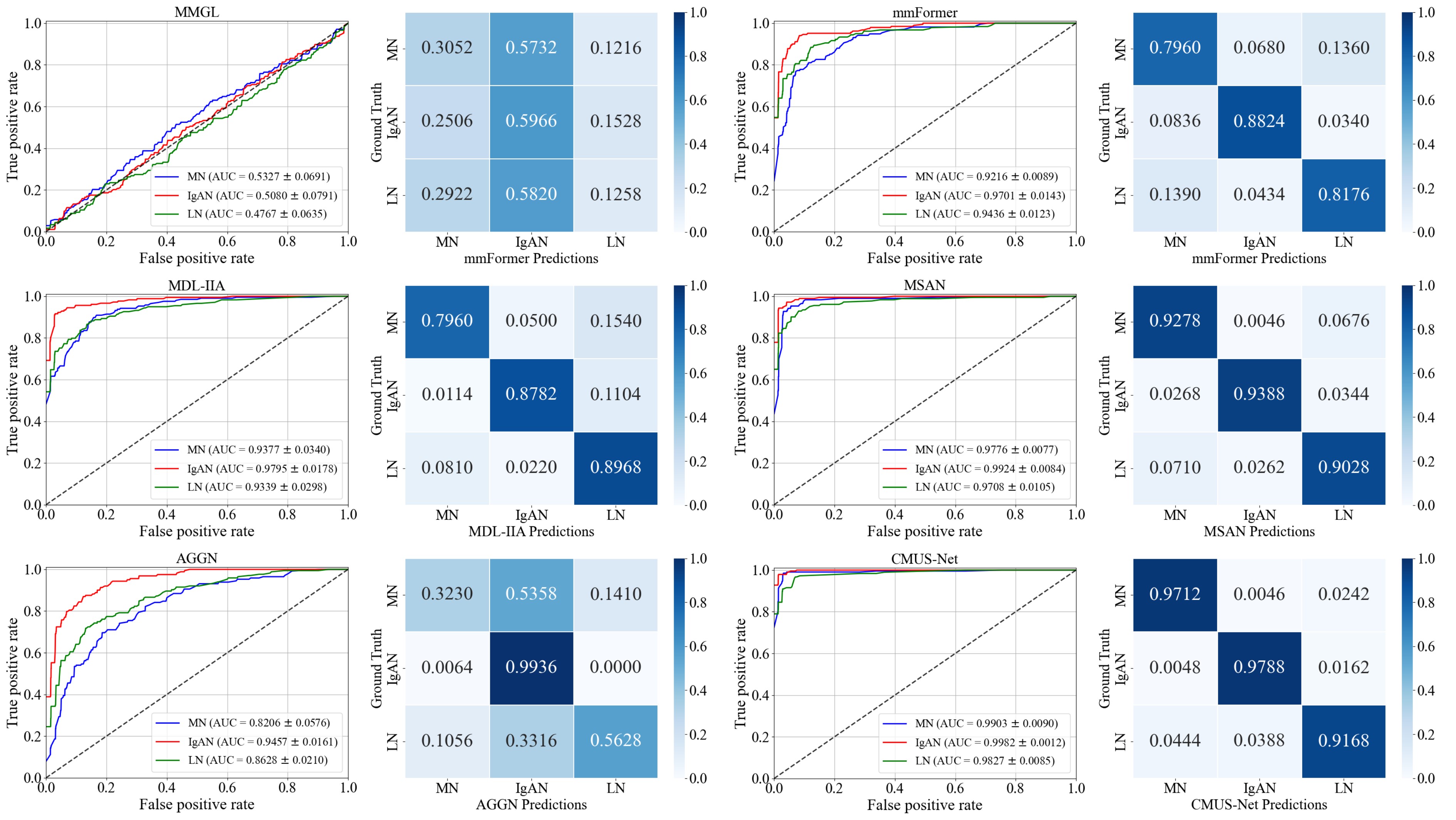}
	\caption{ROCs and confusion matrices of multi-modal models.}
	\label{fig7}
\end{figure*}

We compared our model with other multi-modal models, well-known unimodal classification models, and unimodal multi-scale models. The results for each model on the multi-disease classification dataset are presented in Table~\ref{table4}, with their operating characteristic curves (ROCs) and confusion matrices summarized in Fig.~\ref{fig7}. Our proposed CMUS-Net achieve the best performance with an ACC of 95.37$\pm$2.41\%, an AUC of 99.05$\pm$0.53\%, and an F1 score of 95.32$\pm$2.41\%.

\begin{figure*}[thb]
	\centering
	\includegraphics[width=\textwidth]{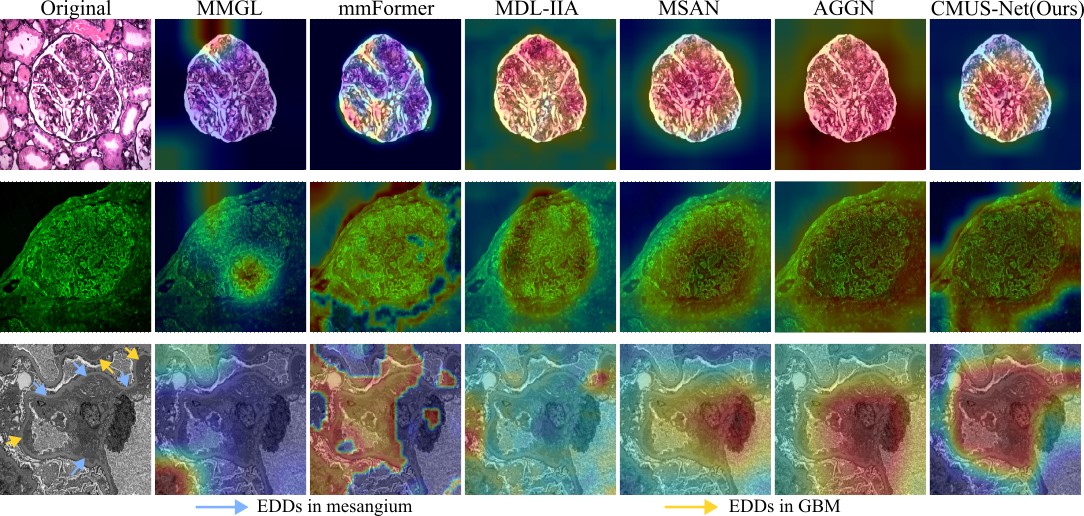}
	\caption{Visualization of class activation maps.}
	\label{fig8}
\end{figure*}

Class activation maps (CAMs) visually emphasize the areas within an image that influence a model’s decision-making process, thereby highlighting its effectiveness in detecting lesions. Fig.~\ref{fig8} illustrates the CAMs generated by our model across three modalities of patients with LN. The lesions in this diseases are diffusely distributed within the glomeruli. As shown in Fig.~\ref{fig8}, our model focuses on the glomerular morphological structure in OM images and the locations of immune-complex deposits in IM images. In TEM images, the model pays attention to the electron-dense deposits across both the mesangium and GBM of LN.

Secondly, we compare our model with well-known unimodal classification models including VGG16 \citep{Simonyan2014VeryDC}, ResNet-50 \citep{He2015DeepRL}, DenseNet-121 \citep{Huang2016DenselyCC}, and ViT-B \citep{Dosovitskiy2020AnII} as well as unimodal multi-scale models such as FPN \citep{Lin2016FeaturePN}, InceptionV3 \citep{xia2017inception}, Res2Net-50 \citep{gao2019res2net}, SwinT-B \citep{Liu2021SwinTH}, CrossViT \citep{Chen2021CrossViTCM}, CrossFormer++ \citep{wang2023crossformer++}, and HiFuse \citep{huo2024hifuse}. To adapt the FPN, originally designed for object detection, to our classification tasks, we replace its detection head with a classification head consisting of two fully connected layers. Each model extracts features of OM, IM, and TEM modalities, respectively, and concatenates them together for classification. Their classification results are shown in Table~\ref{table5}. Our model outperforms other models across all metrics, attributed to the CMUS-Net framework facilitating feature interaction between micrometer- and nanometer-scale images. This enables the model to capture the relationships between pathological changes on OM, IM, and TEM images across scales, leading to superior performance.

\begin{table*}[htb]
	\caption{\label{table4}Comparison results with other multi-modal models on multi-disease classification dataset. \textbf{Bold} denotes the best result in each column. The \textit{\textbf{p}}-value displays the paired t-test results compared to the AUC value of the reference (\textbf{Ref.}).}
	\centering
	\small
	\resizebox{\linewidth}{!}{
		\begin{tabular}{@{}l l *{7}{c}@{}} % use tabularx
			\toprule 
			Types & Models   & ACC(\%)        & AUC(\%)        & PRE(\%)        & REC(\%)        &SPE(\%)         & F1-score(\%)   & $p$-value \\
			\midrule
			\multirow{3}{*}{Multi-modal} 
			      & MMGL     & 36.04$\pm$5.33 & 50.59$\pm$5.72 & 24.98$\pm$6.00 & 34.25$\pm$4.22 & 67.13$\pm$33.0 & 25.46$\pm$5.13 & 0.000 \\
			      & mmFormer & 82.96$\pm$3.49 & 94.52$\pm$0.81 & 83.23$\pm$3.11 & 83.20$\pm$3.10 & 91.60$\pm$4.25 & 82.29$\pm$3.25 & 0.000 \\
			      & MDL-IIA  & 85.74$\pm$5.02 & 95.04$\pm$2.04 & 87.35$\pm$3.99 & 85.70$\pm$5.03 & 92.85$\pm$7.24 & 85.64$\pm$4.80 & 0.000 \\
			\midrule
			\multirow{2}{*}{Multi-modal multi-scale}
			      & MSAN     & 92.22$\pm$0.94 & 98.03$\pm$0.53 & 92.77$\pm$0.77 & 92.33$\pm$0.99 & 96.16$\pm$2.44 & 92.17$\pm$0.90 & 0.003 \\
			      & AGGN     & 63.13$\pm$11.2 & 87.64$\pm$3.00 & 66.74$\pm$12.8 & 62.65$\pm$11.6 & 81.33$\pm$23.5 & 58.88$\pm$14.6 & 0.000 \\
			\midrule
			\multirow{1}{*}{Cross-modal scale} 
			      & CMUS-Net(Ours)     & \textbf{95.37$\pm$2.41} & \textbf{99.05$\pm$0.53} & \textbf{95.29$\pm$2.46} & \textbf{95.52$\pm$2.29} & \textbf{97.78$\pm$1.67} & \textbf{95.32$\pm$2.41} & \textbf{Ref.} \\
			\bottomrule
		\end{tabular}
	%}
}
\end{table*}

\begin{table*}[htb]
	\caption{\label{table5}Comparison results with well-known unimodal classification models and unimodal multi-scale models on multi-disease classification dataset. \textbf{Bold} denotes the best result in each column. The \textit{\textbf{p}}-value displays the paired t-test results compared to the AUC value of the reference (\textbf{Ref.}).}
	\centering
	\small
	\resizebox{\linewidth}{!}{
	\begin{tabular}{@{}l l *{7}{c}@{}}
		\toprule 
		Types & Models               & \makecell{ACC\\(\%)} & \makecell{AUC\\(\%)} & \makecell{PRE\\(\%)} & \makecell{REC\\(\%)} & \makecell{SPE\\(\%)} & \makecell{F1-score\\(\%)} & $p$-value \\
		\midrule
		\multirow{4}{*}{Unimodal} 
		& VGG16                & 90.93$\pm$1.48 & 97.83$\pm$0.83 & 90.87$\pm$1.31 & 91.15$\pm$1.20 & 95.57$\pm$2.70 & 90.86$\pm$1.36 & 0.005 \\
		& ResNet-50            & 76.85$\pm$4.61 & 94.17$\pm$2.19 & 81.47$\pm$2.91 & 76.67$\pm$4.71 & 88.33$\pm$11.5 & 76.16$\pm$4.47 & 0.001 \\
		& DenseNet-121         & 90.37$\pm$0.74 & 97.64$\pm$0.49 & 90.32$\pm$0.56 & 90.46$\pm$0.61 & 89.84$\pm$1.39 & 90.27$\pm$0.61 & 0.000 \\
		& ViT-B                & 71.48$\pm$6.96 & 86.86$\pm$5.29 & 71.80$\pm$6.87 & 71.49$\pm$6.36 & 85.74$\pm$6.07 & 71.35$\pm$6.61 & 0.000 \\
		\midrule
		\multirow{7}{*}{Unimodal multi-scale} 
		& FPN                  & 42.41$\pm$3.95 & 69.37$\pm$3.32 & 51.33$\pm$17.0 & 41.35$\pm$4.11 & 49.42$\pm$15.0 & 34.41$\pm$8.51 & 0.000 \\
		& InceptionV3          & 41.67$\pm$6.80 & 63.95$\pm$4.33 & 40.11$\pm$4.43 & 42.61$\pm$4.70 & 71.31$\pm$29.4 & 34.85$\pm$7.92 & 0.000 \\
		& Res2Net-50           & 90.74$\pm$1.94 & 97.09$\pm$0.51 & 90.70$\pm$2.00 & 90.83$\pm$1.82 & 85.00$\pm$5.34 & 90.59$\pm$1.86 & 0.000 \\
		& SwinT-B              & 91.30$\pm$1.72 & 97.77$\pm$0.69 & 91.33$\pm$1.27 & 91.59$\pm$1.28 & 95.80$\pm$2.89 & 91.21$\pm$1.50 & 0.000 \\
		& CrossViT             & 89.63$\pm$2.14 & 97.58$\pm$0.38 & 89.69$\pm$1.97 & 89.81$\pm$2.29 & 91.60$\pm$1.27 & 89.55$\pm$2.14 & 0.001 \\
		& CrossFormer++        & 63.70$\pm$11.2 & 79.64$\pm$8.95 & 65.39$\pm$11.5 & 64.53$\pm$10.9 & 82.26$\pm$11.9 & 62.90$\pm$12.7 & 0.000 \\
		& HiFuse               & 41.48$\pm$6.56 & 75.49$\pm$4.43 & 39.55$\pm$20.4 & 42.67$\pm$7.42 & 60.01$\pm$6.99 & 31.10$\pm$9.99 & 0.000 \\
		\midrule
		\multirow{1}{*}{Cross-modal scale} 
		& CMUS-Net(Ours)       & \textbf{95.37$\pm$2.41} & \textbf{99.05$\pm$0.53} & \textbf{95.29$\pm$2.46} & \textbf{95.52$\pm$2.29} & \textbf{97.78$\pm$1.67} & \textbf{95.32$\pm$2.41} & \textbf{Ref.} \\
		\bottomrule
	\end{tabular}
}
\end{table*}

\begin{table*}[htb]
	\caption{\label{table6}Comparison results with other models on MN staging dataset. \textbf{Bold} denotes the best result in each column. The \textit{\textbf{p}}-value displays the paired t-test results compared to the AUC value of the reference (\textbf{Ref.}).}
	\centering
	\small
	\resizebox{\linewidth}{!}{
	\begin{tabular}{@{}l l *{7}{c}@{}}
		\toprule 
		Types & Models               & \makecell{ACC\\(\%)} & \makecell{AUC\\(\%)} & \makecell{PRE\\(\%)} & \makecell{REC\\(\%)} & \makecell{SPE\\(\%)} & \makecell{F1-score\\(\%)} & $p$-value \\
		\midrule
		\multirow{4}{*}{Unimodal} 
		& VGG16                & 65.78$\pm$3.31 & 80.80$\pm$3.00 & 72.60$\pm$11.7 & 62.57$\pm$3.00 & 65.16$\pm$6.69 & 65.95$\pm$3.24 & 0.002 \\
		& ResNet-50            & 55.38$\pm$7.34 & 68.83$\pm$3.61 & 44.28$\pm$17.5 & 44.33$\pm$6.64 & 56.63$\pm$17.0 & 39.02$\pm$10.0 & 0.000 \\
		& DenseNet-121         & 68.08$\pm$4.14 & 81.44$\pm$1.99 & 66.34$\pm$5.01 & 66.63$\pm$3.72 & 68.12$\pm$10.9 & 65.40$\pm$3.78 & 0.001 \\
		& ViT-B                & 52.69$\pm$6.51 & 69.92$\pm$3.97 & 51.00$\pm$12.7 & 49.34$\pm$5.31 & 64.38$\pm$4.39 & 47.37$\pm$7.37 & 0.000 \\
		\midrule
		\multirow{7}{*}{Unimodal multi-scale} 
		& FPN                  & 50.00$\pm$10.1 & 57.94$\pm$4.20 & 36.45$\pm$13.8 & 37.25$\pm$4.91 & 54.44$\pm$26.8 & 33.16$\pm$4.95 & 0.000 \\
		& InceptionV3          & 55.77$\pm$4.39 & 64.49$\pm$5.50 & 47.11$\pm$14.1 & 42.03$\pm$5.09 & 53.40$\pm$7.75 & 36.22$\pm$3.33 & 0.000 \\
		& Res2Net-50           & 63.46$\pm$4.39 & 79.13$\pm$2.06 & 62.69$\pm$4.94 & 61.02$\pm$3.11 & 58.45$\pm$8.81 & 59.92$\pm$2.45 & 0.003 \\
		& SwinT-B              & 60.00$\pm$5.63 & 79.55$\pm$4.58 & 58.13$\pm$10.2 & 55.20$\pm$5.41 & 68.40$\pm$4.73 & 54.77$\pm$6.65 & 0.001 \\
		& CrossViT             & 65.77$\pm$4.11 & 82.45$\pm$4.90 & 65.99$\pm$9.59 & 59.38$\pm$5.52 & 70.83$\pm$4.10 & 60.01$\pm$6.80 & 0.007 \\
		& CrossFormer++        & 54.23$\pm$2.55 & 67.07$\pm$3.17 & 42.36$\pm$6.06 & 45.83$\pm$4.09 & 70.33$\pm$11.7 & 42.61$\pm$5.40 & 0.000 \\
		& HiFuse               & 42.31$\pm$10.0 & 69.72$\pm$6.05 & 36.78$\pm$20.0 & 40.89$\pm$7.42 & 66.65$\pm$9.99 & 31.60$\pm$10.2 & 0.000 \\
		\midrule
		\multirow{3}{*}{Multi-modal}
		& MMGL                 & 35.66$\pm$15.1 & 47.07$\pm$1.75 & 19.81$\pm$3.91 & 28.87$\pm$4.08 & 66.09$\pm$1.98 & 21.33$\pm$4.88 & 0.000 \\
		& mmFormer             & 56.92$\pm$5.25 & 67.80$\pm$4.88 & 48.34$\pm$8.24 & 50.42$\pm$6.02 & 88.68$\pm$3.91 & 47.53$\pm$5.45 & 0.000 \\
		& MDL-IIA              & 61.15$\pm$5.36 & 77.03$\pm$2.33 & 63.87$\pm$4.60 & 53.80$\pm$65.9 & 57.86$\pm$10.2 & 52.71$\pm$8.36 & 0.000 \\
		\midrule
		\multirow{2}{*}{Multi-modal multi-scale}
		& MSAN                 & 59.62$\pm$8.07 & 73.46$\pm$3.32 & 55.54$\pm$14.5 & 55.70$\pm$7.36 & 70.76$\pm$3.88 & 52.71$\pm$8.09 & 0.001 \\
		& AGGN                 & 52.50$\pm$3.64 & 64.07$\pm$4.69 & 37.08$\pm$19.2 & 39.60$\pm$6.29 & 65.63$\pm$5.20 & 34.01$\pm$9.60 & 0.000 \\
		\midrule
		\multirow{1}{*}{Cross-modal scale}
		& CMUS-Net(Ours)       & \textbf{72.69$\pm$3.73} & \textbf{88.29$\pm$3.61} & \textbf{74.17$\pm$3.96} & \textbf{69.21$\pm$5.64} & \textbf{84.60$\pm$2.82} & \textbf{69.61$\pm$4.92} & \textbf{Ref.} \\
		\bottomrule
	\end{tabular}
}
\end{table*}

\subsection{Comparison results on MN staging dataset}
\label{section5-3}
\begin{figure}[htb]
	\centering
	\includegraphics[width=\columnwidth]{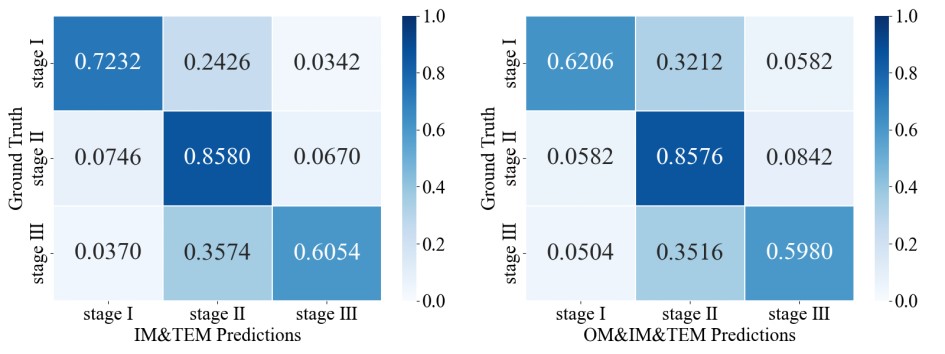}
	\caption{The confusion matrices of CMUS-Net when input with IM and TEM modalities versus input with OM, IM, and TEM modalities.}
	\label{fig9}
\end{figure}

The proposed method excels not only in the glomerular multi-disease classification but also extends its utility to other classification tasks in the renal biopsy process, such as MN staging. We compare it with other models on the MN staging dataset, with performance metrics summarized in Table~\ref{table6}. CMUS-Net outperforms other models with mean ACC of 72.69\%, mean AUC of 88.29\%, and mean F1-score of 69.61\%, which suggests that our proposed method has a certain generalizability in renal biopsy image analysis. Besides, we compared the classification performance of CMUS-Net with different modality combinations as inputs on the MN staging dataset, and the results are shown in Table~\ref{table7}. Fig.~\ref{fig9} shows the confusion matrices of the optimal and suboptimal results in Table~\ref{table7}.

\begin{table*}[tb]
	\caption{\label{table7}Comparison results of different modal combinations on MN staging dataset. \textbf{Bold} denotes the best result in each column. The \textit{\textbf{p}}-value displays the paired t-test results compared to the AUC value of the reference (\textbf{Ref.}).}
	\centering
	\small
	\resizebox{\linewidth}{!}{
		\begin{tabular}{@{}l l *{8}{c}@{}}
			\toprule 
			  OM         & IM         & TEM        & ACC(\%)                 & AUC(\%)                    & PRE(\%)                 & REC(\%)                 & SPE(\%)                 & F1-score(\%)   & $p$-value \\
			\midrule 
			  \checkmark &            &            & 66.59$\pm$4.03          & 82.77$\pm$1.87             & 66.53$\pm$2.92          & 63.60$\pm$4.73          & 81.81$\pm$13.6          & 62.63$\pm$3.88 & 0.000 \\
			             & \checkmark &            & 58.35$\pm$8.14          & 73.07$\pm$5.66             & 55.87$\pm$11.8          & 53.00$\pm$8.25          & 76.47$\pm$16.5          & 50.98$\pm$9.93 & 0.000 \\
			             &            & \checkmark & 71.46$\pm$7.54          & 87.31$\pm$4.10 & 72.40$\pm$9.06          & 69.47$\pm$8.47          & 84.74$\pm$14.5          & 69.51$\pm$8.34 & 0.346 \\
			  \checkmark & \checkmark &            & 66.54$\pm$1.54          & 78.19$\pm$2.78             & 66.12$\pm$4.95          & 64.45$\pm$0.83          & 82.22$\pm$11.5          & 63.83$\pm$1.48 & 0.000 \\
			  \checkmark &            & \checkmark & 73.08$\pm$2.43          & 86.35$\pm$2.13             & 71.99$\pm$7.30          & 68.11$\pm$5.70          & 84.06$\pm$13.8          & 68.69$\pm$6.71 & 0.006 \\
			             & \checkmark & \checkmark & \textbf{75.39$\pm$4.93} & \textbf{88.99$\pm$2.90}    & \textbf{76.93$\pm$5.29} & \textbf{72.93$\pm$5.02} & \textbf{85.68$\pm$14.3} & \textbf{73.73$\pm$5.02} & \textbf{Ref.} \\
			  \checkmark & \checkmark & \checkmark & 72.69$\pm$3.73          & 88.29$\pm$3.61 & 74.17$\pm$3.96          & 69.21$\pm$5.64          & 84.60$\pm$2.82          & 69.61$\pm$4.92 & 0.684 \\
			\bottomrule
		\end{tabular}
}
\end{table*}

\section{Discussion}
\label{section6}
We propose a novel approach, CMUS-Net, as a second opinion system to achieve glomerular multi-disease classification by integrating OM, IM, and TEM images. Current methods face challenges in effectively fusing modal features due to the significant scale differences between nanometer-scale TEM images and micrometer-scale OM and IM images. Our CMUS-Net aims to overcome these challenges by incorporating a SMIL module to aggregate information from multiple TEM images to bridge the scale difference between modalities, a CMSA module to enhance the relevant pathological semantic information across modalities, a weighted loss function to strengthen the model’s awareness of the varying importance of different modalities.

Our proposed SMIL and CMSA modules effectively promote the feature interaction process. Fig.~\ref{fig3} shows the result compared to other attention mechanisms. Contrary to expectations, both the Self-Attention and the Cross-modal Attention model perform worse than the Without Attention model. This can be attributed to two key factors. Firstly, self-attention focuses primarily on features within modalities, which may either ignore or exacerbate scale differences among features. Secondly, while cross-modal attention facilitates feature interaction between pairs of modalities, it lacks explicit guidance from scale relationships, leading to suboptimal performance.

The CMUS-Net framework is able to improve the encoder’s capability to extract and distinguish features of multiple diseases, achieving an effective fusion of cross-modal scale features. Fig.~\ref{fig4} shows the clustering results of the encoders. Comparing the first and second rows of , we observe that the classification ACC of the multi-modal model (91.30$\pm$1.72\%) does not surpass that of the best-performing unimodal model (91.30$\pm$1.26\%). This is because multi-modal feature fusion from renal biopsy images is not a straightforward process where “one plus one equals two” \citep{Zhang2023MultimodalRL}. The massive scale differences between micrometer and nanometer images hinder the multi-modal models from effectively learning and integrating features from each modality, resulting in the encoders failing to utilize these features for disease classification. When comparing the last row with the first two rows of Fig.~\ref{fig4}, it is evident that our method significantly enhances the feature extraction and discrimination capabilities of the encoders, particularly for the IM and TEM modalities. This improvement can be attributed to CMUS-Net’s ability to guide each encoder in capturing and enhancing relevant pathological features across different modalities by leveraging the scale relationships between renal biopsy images. While some encoders’ performance still falls short of that in unimodal models, our proposed method effectively integrates these features, leading to the model achieving significantly better results (95.37$\pm$2.41\%).

The performance of CMUS-Net varies on the multi-disease classification dataset when input with different modality combinations, as shown in Table~\ref{table3}. Notably, among the three modalities, the model achieves its best with IM input and its worst with TEM input. We attribute this disparity to the distinct imaging methods of these two modalities. IM examination presents the location of immune-complex deposition through specific antibody detection, resulting in images that highlight patterns strongly correlated with immune-mediated glomerular diseases such as IgAN, MN, and LN. Thereby, models can readily utilize the IM modality to achieve satisfactory results. Conversely, TEM examination produces grayscale images of the glomerular ultrastructure with electron beams. The ultrastructures in these images are complex and varied, combined with less pronounced contrasts, making it challenging for the models to extract valuable features. However, these ultrastructural details are integral to the renal biopsy pathological diagnostic system, and the ability to effectively mine and utilize this information is crucial in determining the superiority of model performance. Our proposed CMUS-Net framework aggregates ultrastructure information from TEM images and interacts it with those from OM and IM images, leading to superior model performance.

Our model outperforms other multi-modal models in Table~\ref{table4} across all metrics and demonstrates superior overall performance in classifying the three diseases due to the CMUS-Net framework effectively leveraging the scale relationship between micrometer and nanometer images, guiding the model to integrate complementary features while enhancing relevant semantic features across modalities. MMGL exhibits relatively anomalous performance with an abnormal confusion matrix. This may be attributed to the transform matrices in MMGL being primarily used for extracting simple features, which are less effective at capturing the complex features present in renal biopsy images. Compared to MSAN, another multi-modal multi-scale approach, AGGN additionally integrates multi-scale features across modalities. However, AGGN shows inferior performance in classifying MN and LN, with its overall performance even falling short of two multi-modal single-scale models, mmFormer and MDL-IIA. This suggests that if the scale differences between modalities are too huge, directly fusing multi-scale features from different modalities may not yield optimal results. Nevertheless, we acknowledge that although the compared models were fine-tuned to maximize their performance on renal biopsy image classification, they might still be at a disadvantage due to modality transformation.

The CMUS-Net framework excels not only in the glomerular multi-disease classification but is also applicable to other renal biopsy image classification tasks, such as MN staging, as suggested by Table~\ref{table6}. Surprisingly, on the MN staging dataset, our method performs better with just the IM and TEM modalities as input compared to utilizing all three modalities, as shown in Table~\ref{table7}. We speculate that this may be attributed to the characteristics of early-stage MN: while IM examination is most sensitive to immune complexes and can indicate their approximate deposition location and TEM examination can observe the size, appearance, and specific locations of the deposits, whereas OM examination may only reveal an essentially normal appearance of glomeruli \citep{lerner2021conceptual}. Our method captures and enhances these pathological features, paradoxically resulting in misjudgments by the CMUS-Net. Fig.~\ref{fig9} shows the confusion matrices of CMUS-Net using IM and TEM modalities as inputs, as well as utilizing OM, IM, and TEM modalities as inputs, as detailed in Table~\ref{table7}. Compared with IM and TEM modalities, CMUS-Net performs less accurately in classifying stage \uppercase\expandafter{\romannumeral1} of MN with all three modalities, corroborating our previous speculation to some extent. Furthermore, the performance of multi-modal models on the MN staging dataset is lower than that on the disease classification dataset. This discrepancy arises because, although each stage of MN exhibits distinct pathological characteristics and has well-defined diagnostic criteria, there are numerous similar features across stages, such as electron-dense deposits in the basement membrane and diffuse fusion of podocyte foot processes, which are common in MN. To achieve precise MN staging, models need to filter out the interference from these confounding features. Therefore, the MN staging task is more challenging.

Currently, most images in multi-modal public datasets are at the tissue level \citep{ruckert2024rocov2,Siragusa2024MedPix2A,menze2014multimodal}, and public datasets containing cross-scale images remain scarce. This scarcity poses a challenge for comparing our method with others. Moreover, due to the difficulty of data acquisition, we only evaluated the multi-disease classification performance of CMUS-Net for three common glomerular diseases—MN, IgAN, and LN—without exploring others. Consequently, the model may be biased toward the morphological and ultrastructural patterns specific to these three target classes. Regarding MN staging, when sufficient data for stage MN-IV is available, CMUS-Net can be retrained simply by replacing its final classification layer.

Despite its excellent performance in glomerular multi-disease classification and MN staging tasks, our CMUS-Net has certain limitations. For instance, due to the lesion characteristics of early-stage MN, the performance of CMUS-Net, which takes three modalities as input simultaneously, is suboptimal. To address this issue, it is worth considering to dynamically adjust the importance of each modality so that the model can adapt to a broader range of renal biopsy classification tasks. Additionally, in the current workflow, OM images need to undergo semi-automatic segmentation before being fed into CMUS-Net, which may limit the flexibility of CMUS-Net in practical applications. This could be improved by incorporating an automatic localization and segmentation network to effectively separate the glomerular regions from the background in OM or IM images.

\section{Conclusion}
\label{section7}
In this paper, we propose the CMUS-Net framework for glomerular multi-disease classification. Unlike existing methods, we integrate OM, IM, and TEM images, introducing the SMIL module to bridge the scale differences between micrometer and nanometer scales, employing the CMSA module to enhance pathological semantic information, and utilizing a weighted loss function to boost the model’s awareness of modality importance, thereby improving classification performance. We conducted extensive experiments on datasets for glomerular multi-disease classification and MN staging, and the results demonstrate that our proposed CMUS-Net achieves outstanding performance among classification methods.

\section*{Acknowledgments}
This work was supported by a grant from the National Natural Science Foundation of China (No. 32071368).

\bibliographystyle{plainnat}
\bibliography{refs.bib}

\end{document}